\documentclass[authoryear,preprint,12pt]{elsarticle}

\usepackage{graphicx}%
\usepackage{multirow}%
\usepackage{amsmath,amssymb,amsfonts}%
\usepackage{amsthm}%
\usepackage{mathrsfs}%
\usepackage[title]{appendix}%
\usepackage{xcolor}%
\usepackage{textcomp}%
\usepackage{manyfoot}%
\usepackage{booktabs}%
\usepackage{algorithm}%
\usepackage{algorithmicx}%
\usepackage{algpseudocode}%
\usepackage{listings}%
\usepackage{graphicx}

\usepackage{stfloats}
\usepackage{booktabs}
\usepackage{fancyvrb}
\usepackage{makecell}

\usepackage{adjustbox}
\usepackage{pdflscape}
\usepackage{booktabs}
\usepackage{array}
\usepackage{url}
\usepackage{hyperref}

\begin{document}

\begin{frontmatter}



\title{WrAFT: a Modularized Automated Writing Evaluation System for Argumentative Essays}


\author[inst1]{Adnan Labib}

\author[inst2]{Yixuan Huang}

\author[inst3]{Jiahui Wu}

\author[inst2]{John Maurice Gayed}

\author[inst4]{Zheng Yuan}

\author[inst5]{Qiao Wang\corref{cor1}}
\ead{judy.wang@hosei.ac.jp}
\cortext[cor1]{Corresponding author}

\affiliation[inst1]{organization={Department of Informatics, King's College London},
            city={London},
            country={UK}}

\affiliation[inst2]{organization={Waseda University},
            city={Tokyo},
            country={Japan}}

\affiliation[inst3]{organization={Beijing University of Posts and Telecommunications},
            city={Beijing},
            country={China}}

\affiliation[inst4]{organization={University of Sheffield},
            city={Sheffield},
            country={UK}}

\affiliation[inst5]{organization={Hosei University},
            city={Tokyo},
            country={Japan}}

\begin{abstract}
This study presents \textit{WrAFT}, a \textbf{Wr}iting \textbf{A}ssessment and \textbf{F}eedback \textbf{T}ool, that delivers both accurate and reliable scores and effective comprehensive feedback to argumentative essays. \textit{WrAFT} adopts a modular design by dividing the automated writing evaluation (AWE) tasks into scoring, surface-level feedback and deep-level feedback. In building the system, various Large Language Models (LLMs) have been evaluated, including LLaMA-3.3-70B-Instruct, GPT-4o, and Claude 3.7, through both direct prompting and supervised fine-tuning approaches. A proprietary dataset of 480 TOEFL Independent Writing essays with official benchmark scores was utilized. Benchmark-based evaluation shows that \textit{WrAFT} achieves state-of-the-art performance in scoring with a quadratic weighted kappa (QWK) of 0.84 and an root mean square error (RMSE) of 0.44 against official scores on a scale of 0-5. Human evaluation of system-generated feedback also reveals high approval ratings (96.14\% for surface-level, 93.03\% for deep-level macro feedback, and 94.69\% for deep-level micro feedback). An interactive user interface has been developed for the system, publicly available and free to use.
\end{abstract}






\end{frontmatter}



\section{Introduction}\label{sec1}

Argumentative writing is emphasized in secondary and post-secondary curricula as it fosters higher-order thinking \cite{graff2003, kuhn2005}. Assessing such open-ended writing reliably, however, is a notoriously difficult task for human raters: it demands considerable time, and even trained evaluators can exhibit subjective biases and inconsistency in their judgments \citep{shermis2003}. Automated Writing Evaluation (AWE) systems have thus emerged as a promising solution to score and provide feedback on student essays at scale.

Traditional AWE systems have mostly focused on scoring \citep{li-ng-2024-automated}, either through linguistic features, shallow text similarity measures \citep{li2023automated} or through neural network approaches to model input essays, giving grades based on a single vector representation of the essay \citep{dong-etal-2017-attention, jin-etal-2018-tdnn}. Though some of the systems demonstrated satisfactory scoring agreements with human raters, such as \textit{e-rater}\footnote{\url{https://www.ets.org/erater/about.html}}, the scoring engine of \textit{Criterion}, Education Testing Service's (ETS) AWE tool for the GRE and TOEFL writing exams \citep{attali2006}, they often failed to capture higher-order thinking in writing such as coherence or argumentation. 

The emergence of powerful Large Language Models (LLMs) opens the door to AWE systems that not only predict a score but also give rich explanatory feedback on content. 
However, a comprehensive review of AWE research by \citet{li-ng-2024-automated} shows that in recent years, this area of research seems to be narrowly focused on developing a sophisticated model that can beat competing models in scoring a standard evaluation dataset, such as the Automated Student Assessment Prize (ASAP\footnote{\url{https://www.kaggle.com/datasets/lburleigh/asap-2-0}}) and the Cambridge Learner Corpus-First Certificate in English exam (CLCFCE; \citeauthor{yannakoudakis-etal-2011-new}), while there is a lack feedback generation and validation. \citet{li-ng-2024-automated} proposes that there should be different layers to AWE systems, including holistic or trait-specific scores, written feedback, and essays revised by experts. This proposal coincides with the need for comprehensive feedback stressed by second language writing research, which typically includes both corrective edits to surface-level errors in grammar, wording and mechanics, and deep-level feedback comments on issues such as organization, coherence and argumentation \citep{bitchener2008value}. While some researchers caution against the potential cognitive overload that comprehensive feedback might impose on learners \citep{truscott1996case}, comprehensive feedback remains a prevalent pedagogical practice in language education, as learners may selectively attend to aspects they find most relevant. However, though there have been many grammar error correction (GEC) systems that can be used for surface-level feedback \citep{yuan-briscoe-2016-grammatical,imamura-etal-2012-grammar, rozovskaya-roth-2019-grammar, davis-etal-2024-prompting,10.1162/coli_a_00478,yuan-etal-2021-multi}, deep-level feedback remains rare and is usually in a generic manner (e.g., only pointing out that ``the essay could use a clearer thesis and more examples'') \citep{liu2015investigating, ranalli2021l2}.  
In addition, feedback comments are frequently consolidated in standalone paragraphs separate from the essay text, as observed in \cite{stahl2024}, which makes it inconvenient for users to associate the feedback with specific elements of their writing. 

One notable AWE system developed most recently, AiAWE \citep{gayed2025aiawe}, provides highly reliable and accurate TOEFL-benchmarked scoring and holistic feedback to argumentative essays through LLMs finetuned on a proprietary TOEFL independent writing dataset provided by Education Testing Service (ETS). Inspired by the AiAWE system and drawing on the first two layers of feedback proposed by \citet{li-ng-2024-automated}, this research aims to address the current research gap by developing and validating an LLM-based AWE system for argumentative essays that provides accurate and reliable scoring and layered, comprehensive feedback through an interactive user interface (UI). Specifically, a modular architecture is adopted with separate modules for scoring, surface-level feedback and deep-level feedback. A proprietary dataset from ETS consisting of 480 TOEFL Independent Writing essays (the same as used in AiAWE) with benchmark scores on the original scale of 0-5 is used. 
The contribution of this research is as follows:

\begin{itemize}
  \item The scoring module of the system achieves state-of-the-art (SOTA) performance with a QWK of \textbf{0.84} and an RMSE of \textbf{0.44} on a 0-5 score scale.
  \item Human validation of the comprehensive feedback generated by the AWE system suggests highly satisfactory performance in correcting surface-level grammatical and mechanical errors, as well as in commenting on macro structure and micro aspects including context-dependent grammar, clarity, coherence, argumentation, and formality.
  \item Methodology-wise, the research shows that for feedback generation, a task with non-deterministic output, supervised fine-tuning was less effective than directly prompting SOTA LLMs to elicit feedback on specific traits/aspects in writing. 
   \item The research results in a deployable code package, available at \url{https://github.com/judywq/wraft} and a web-based application for free of use with an interactive UI to visualize comprehensive feedback through in-line corrective edits and anchored comments, available at \url{https://wraft.net/}.
\end{itemize}

\section{Background}

\subsection{Early automated writing evaluation}

Automated writing evaluation (AWE) is defined as ``the process of evaluating and scoring written prose via computer programs'' \cite[p.1]{shermis2024introduction}, which includes the dual tasks of scoring and feedback. AWE has a long history in NLP and education research, dating back to the seminal work of \citet{page1967grading} who first outlined the possibility of grading essays by computer and developed Project Essay Grade \citep{page2003project} for automated assessment. Early AWE systems focused primarily on essay scoring. Some of these systems adopt machine learning approaches that either target specific textual features, such as grammar or lexical sophistication, or focus on semantic similarity using techniques such as latent semantic analysis. For instance, 
\textit{\textit{e-rater}} \citep{attali2006} from ETS utilizes a suite of handcrafted linguistic features, such as grammar errors, vocabulary sophistication, organization indicators, etc. to evaluate writing. The Intelligent Essay Assessor (IEA) \citep{landauer2003} employs latent semantic analysis to measure the semantic similarity between an essay and high-scoring responses. More recent AWE systems have adopted deep learning techniques, in which models learn distributed representations of essays such that texts of similar quality are mapped to similar vector spaces \citep{li-ng-2024-automated}. One example is the work of \citet{taghipour-ng-2016-neural}, who employed a convolutional neural network (CNN) to extract n-gram-level features for capturing local dependencies, followed by a long short-term memory network (LSTM) to model global, long-distance dependencies for holistic essay scoring. \citet{shermis2014state} provided a comprehensive evaluation of AWE scoring in high-stakes contexts. Including large-scale validation studies and an international AWE competition, the researcher demonstrated AWE systems could approach human scoring performance across multiple dimensions. While Shermis emphasized the operational potential of these systems, later critiques (see \citeauthor{perelman2014state}, \citeyear{perelman2014state}) show that many of these scoring engines disproportionately reward surface-level features, especially essay length (word count), rather than substantive content, raising questions about the depth of construct representation. 

Further investigation shows that while some of the above systems have proven effective for scoring, such as \textit{e-rater} \citep{burstein2004automated}, a common limitation is their lack of true understanding of content, particularly higher-order thinking such as logic and argumentation. Thus, feedback in early AWE system, if any, tends to be formulaic (e.g., pointing out grammatical and mechanical errors) rather than giving insight into argument strength or coherence \citep{li-ng-2024-automated}. In addition, while \textit{e-rater} has been used extensively in high-stakes environments, \citet{powers2002stumping} demonstrates its weakness in their stress-test of \textit{\textit{e-rater}}, where writing experts and computational linguists were able to deliberately create essays that exploit the \textit{e-rater} algorithm. Their findings highlight early concerns about the validity of algorithm-based AWE, showing that automated systems could reward superficial linguistic features rather than the deeper qualities of coherent and original writing. \citet{deane2013relation} extends on this by examining how AWE scoring aligns with contemporary theories of writing. The researcher argues that while AWE systems can capture certain linguistic and structural features, they risk under-representing higher-order dimensions of writing such as idea development and rhetorical effectiveness. This perspective reinforces the concern that AWE scoring should not be evaluated solely on validity, but needs better alignment with broader construct models of writing proficiency.  

\subsection{LLM-based approaches}

The recent advancements of LLMs has sparked a new wave of research and applications in AWE. With their strong text comprehension and generation capabilities, LLMs can read a student essay and produce a detailed critique in natural language, often well beyond the templated feedback of older AWE systems. \citet{mizumoto2023exploring} were among the first to empirically evaluate the use of LLMs for automated essay scoring. Using GPT-3 to assess essays from ETS's TOEFL11 dataset, the researchers reported the model achieving good levels of agreement and reliability with the human ratings by professional ETS raters. In addition, the researchers noted improvements when combining LLM evaluation with linguistic feature-based approaches. The study demonstrated both the promise and limitations of zero-shot LLM scoring, positioning LLMs as viable, but still needing methodological refinements to ensure consistency and construct validity.
Researchers' efforts to utilize LLMs for AWE tasks include both direct prompting and fine-tuning approaches. In the realm of direct prompting, a study by \citet{mansour2024} evaluated the capabilities of ChatGPT (\href{chatgpt}{https://chatgpt.com/}), the web interface of the GPT models, and LLaMA-2 (\url{https://www.llama.com/llama2/}) in scoring written essays. Through various prompt-engineering tactics, they found that both models exhibited comparable performance in automated essay scoring, with ChatGPT having a slight advantage. Another study by \citet{stahl2024} explored zero-shot and few-shot approaches inspired by Chain-of-Thought prompting \citep{wei2022chain} to generate both scores and feedback. Their study found that addressing both tasks simultaneously, rather than independently, enhances the quality of generated feedback and improves scoring performance, although the impact on feedback quality remains limited. Building on this, \citet{pack2024large} examined the reliability and validity of LLM-based essay scoring for English language learner writing. The researchers' study highlights while LLMs such as GPT models can achieve human-comparable scoring accuracy, performance varied across time and task type, raising questions about reliability and generalizability. Importantly, they highlight that while LLMs offer a significant advancement over earlier rule/feature-based AWE, systemic validity checks remain necessary to ensure LLMs capture more than superficial patterns in student writing. \citet{li2024applying} conducted a large-scale evaluation of applying LLMs for AWE, benchmarking models such as GPT alongside more traditional feature-based approaches. The results indicated LLMs can match or exceed the performance of traditional AWE systems, while also providing more flexible and context-sensitive evaluations of student writing. At the same time, the researchers cautioned that issues of transparency and consistency remain as challenges. 

For studies that involve fine-tuning, researchers have found that in scoring, fine-tuning base models or even older or smaller models has been proven more effective than directly prompting more advanced LLMs \citep{li-ng-2024-automated}. A study by \citet{wang2024} fine-tuned the GPT-3.5 model on a corpus of TOEFL argumentative essays with human benchmark scores and compared fine-tuned models with base GPT-3.5 and GPT-4 models. Their results demonstrated that the fine-tuned models achieved higher scoring accuracy and reliability than zero-shot prompting of GPT-3.5 or GPT-4, and that the fine-tuned models were robust when scoring essays from unseen prompts.  \citet{gayed2025aiawe} builds upon the work of \citet{wang2024} by developing a freely accessible web-based application called AiAWE. The application uses an open-source LLM (Gemma-3-27B-IT) for automated essay scoring by using Low-Rank Adaption (LoRA) adapters and rubric-guided system prompting to score argumentative essays. By preserving Gemma's base-model behavior, the author was able to show both strong agreement with gold-standard ETS raters across standard metrics in addition to providing actionable deep-level rubric-aligned feedback to users. Another study by \citet{cai2025} introduced the Rank-Then-Score (RTS) framework, which employs a two-stage process: first ranking essays using a fine-tuned LLM, then scoring them based on the rankings. This method outperformed traditional supervised fine-tuning techniques, particularly in Chinese datasets. Further integrating fine-tuning and prompt engineering, \citet{chu2024} proposed the Rationale-based Multiple Trait Scoring (RMTS) model. RMTS combines prompt-engineering-based LLMs with a fine-tuning-based essay scoring model to provide trait-specific rationales for scores. Their approach enhances the reliability of multi-trait scoring by generating fine-grained explanations aligned with rubric guidelines.

Despite the recognized potential of LLMs to generate rich, formative feedback, the majority of recent AWE research has continued the traditional focus on scoring accuracy \citep{li-ng-2024-automated}. This persistent emphasis is likely because feedback evaluation is significantly more time-consuming and requires more extensive human labor than scoring. Unlike scores, which can be validated against quantitative benchmarks, the quality of feedback must be assessed through qualitative, pedagogical judgment. As a result, the use of LLMs for feedback remains largely anecdotal rather than proven through rigorous research. The limited number of studies that do cover feedback generation, such as those exploring Chain-of-Thought prompting to produce comments alongside scores \citep{stahl2024} or generating trait-specific rationales \citep{chu2024}, represent important but initial steps. As such, this research aims to develop and validate an AWE system designed to provide both reliable scores and pedagogically valuable, fine-grained feedback. To make the system accessible to end users, an interactive UI will also be built, transforming the system from a research model into a practical and pedagogically effective educational tool.




\section{Methods}



\subsection{Dataset and subsets}
The development of the system in this research required a dataset of argumentative essays with benchmark scores and comprehensive feedback. To this end, a proprietary dataset of 480 TOEFL Independent Writing test-taker samples with official scores from ETS\footnote{Access to this dataset is restricted to researchers approved by ETS under a limited, non-exclusive, revocable, and non-transferable license.} was obtained. TOEFL Independent Writing is a typical argumentation writing task where test-takers write in response to an essay prompt, such as ``\textit{Do you agree or disagree the following statement:...}''
The samples in the dataset are evenly distributed under two essays prompts and two ETS raters delivered integer scores from 0-5 based on specific rubrics\footnote{
\url{https://www.nafsa.org/sites/default/files/ektron/files/underscore/regiii/conference/2009/pruner\%20indepwrihd-t2.pdf} }. Where the discrepancy of the two rater scores was no more than 1, the final score was the average of the two. Otherwise, a third rater was engaged to deliver the final score \citep{TOEFLscore}. As essays scored 0 were excluded, the scores of the essays in the dataset range from 1-5 with 0.5 increments. The score distribution of the 480 essays are shown in \autoref{score_distribution}.

\begin{table}[htbp]
\centering
\caption{Score distribution in the dataset}
\begin{tabular}{llll} \toprule
Score & Prompt 1 & Prompt 2 & Subtotal \\ \midrule
1     & 1        & 2        & 3        \\
1.5   & 3        & 2        & 5        \\
2     & 18       & 20       & 38       \\
2.5   & 32       & 25       & 57       \\
3     & 65       & 55       & 120      \\
3.5   & 40       & 45       & 85       \\
4     & 32       & 38       & 70       \\
4.5   & 35       & 28       & 63       \\
5     & 14       & 25       & 39       \\ \midrule
Total & 240      & 240      & 480      \\ \bottomrule
\end{tabular}
\label{score_distribution}
\end{table}

For the scoring module, 120 essays were selected through score-based equal sampling across the two essay prompts as the fine-tuning subset. As the number of essays scored 1 is limited, essays of other scores were added to make up for it. The remaining 360 essays comprised the test subset.

\begin{figure}[htbp]
    \centering
    \includegraphics[width=0.6\linewidth]{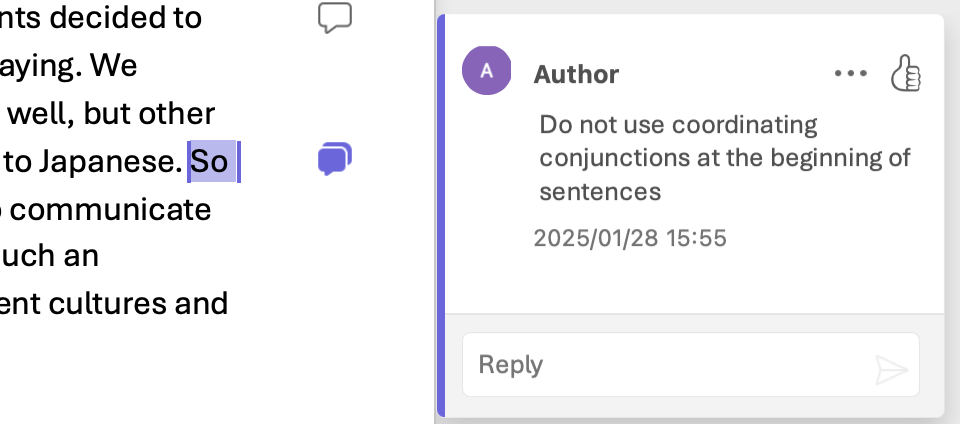}
    \caption{Example of a teacher comment in MS Word}
    \label{MS comment}
\end{figure}

For the feedback module, as no feedback was available in the dataset, this research curated its own datasets for comprehensive feedback. For surface-level feedback in the form of corrective edits to grammar and mechanics, previous research on GEC has shown that directly prompting LLMs generates satisfactory results \citep{davis-etal-2024-prompting}. As such, supervised fine-tuning and the creation of corresponding datasets were not necessary. For deep-level feedback that require higher-order thinking, 90 essays from the 480 essays were selected using score-based equal sampling. Eight experienced university teachers who teach academic writing courses were recruited to provide feedback using Microsoft (MS) Word's comment feature (see~\autoref{MS comment} for an example of the comments). Before annotation, the teachers received training on the scoring rubrics and were given nine benchmark essays scored from 1 to 5 and annotated by the researchers as reference exemplars. The 90 essays had already been processed for surface-level corrections and the scores had been removed to ensure the annotations would focus solely on deep-level features. After the teachers submitted their annotated essays, another independent teacher reviewed all feedback comments to ensure they were accurate and free from surface-level errors. Their feedback data was then processed into JSON format, including the targeted text elements (the start and end character positions of the text string and the tokens in the text element) and the respective comments. This JSON data comprises the fine-tuning subset for deep-level feedback.

For testing of both surface-level and deep-level feedback,
due to the extensive human labor in evaluating such feedback, only 40 essays were selected (through equal sampling) from the 390 non-annotated essays. The summary of datasets used in this study is shown in \autoref{datasets}.

\begin{table*}[htbp]
\centering
\caption{Datasets used in fine-tuning and testing}
\label{datasets}

\begin{adjustbox}{width=\textwidth}
\begin{tabular}{|l|c|l|}
\hline
\textbf{Module - Data Subset} & \textbf{Number of Essays} & \textbf{Data Content} \\
\hline
Scoring - Fine-tune & 120 & Essay prompts, essays, scores, rubrics\\
Scoring - Test & 360 & Essay prompts, essays, scores, rubrics\\\hline
Surface-level - Test & 40 & Essay prompts, essays \\\hline
Deep-level - Fine-tune & 90 & Essay prompts, essays, target text elements, comments \\
Deep-level - Test & 40 & Essay prompts, essays (same with surface-level testing)\\
\hline
\end{tabular}
\end{adjustbox}

\end{table*}

\subsection{LLM choices and procedures}
\paragraph{Scoring LLMs}
For the scoring module, both open-source and commercial models were evaluated to ensure broad applicability. For the open-source option, LLaMA-3.3-70B-Instruct (\url{https://huggingface.co/meta-llama/Llama-3.3-70B-Instruct}) was selected, the top model in instruction following\footnote{\url{https://huggingface.co/spaces/open-llm-leaderboard/open_llm_leaderboard\#/}} at the time this research was conducted.  For the commercial model, following suggestions in \citet{wang2024}, GPT-4o\footnote{\url{https://platform.openai.com/docs/models/gpt-4o}} was selected, as it was the latest model from the GPT series that allows supervised fine-tuning at the time this research took place. The fine-tuning prompt can be found at \autoref{app:scoring prompt}.
\paragraph{Surface-level LLMs}
For the surface-level feedback module based on direct prompting, it is divided into two stages: corrective edits and the generation of explanations for those edits.  In a separate evaluation conducted by the researchers \citep{10.1007/978-3-032-29755-6_41}, DeepSeek-V3-671B\footnote{\url{https://github.com/deepseek-ai/DeepSeek-V3}}, GPT-4o and LLaMA-3.3-70B-Instruct were compared in their GEC performance through zero-shot prompting using the  W\&I+LOCNESS dataset from the BEA 2019 Shared Task \citep{bryant_2019}. Findings (see Figure \ref{tab:baseline-gec-results}) suggest that the three models performed similarly across four commonly used evaluation metrics in GEC, including ERRANT F$_{0.5}$ \citep{bryant_2017}, GLEU \citep{napoles_2016}, PT-ERRANT \citep{gong_2022} and Scribendi \citep{islam-magnani-2021-end}. Drawing on findings from these findings, GPT-4o was chosen for the corrective edits as the small performance margin did not justify the significant infrastructure costs and engineering complexity required to host a large-scale open-source model like the 671B DeepSeek model. To avoid overcorrection, the prompt to GPT4o specifically asked it not to introduce stylistic and word choice corrections. The prompt can be found at \autoref{app:surface-level prompt}.


\begin{table}[htbp]
\centering
\caption{Performance of directly prompted SOTA LLMs on GEC metrics}
\label{tab:baseline-gec-results}
\begin{adjustbox}{width=\textwidth}
\begin{tabular}{lcccc}
\hline
\textbf{Model} & \textbf{ERRANT} & \textbf{GLEU} & \textbf{PT-ERRANT} & \textbf{Scribendi} \\
\hline
DeepSeek-V3-671B & 0.4926 & 0.7677 & 0.5666 & 0.5977 \\
GPT-4o & 0.4592 & 0.7420 & 0.5484 & 0.6843 \\
LLaMA-3.3-70B-Instruct & 0.4826 & 0.7345 & 0.4973 & 0.4710 \\
\hline
\end{tabular}
\end{adjustbox}
\end{table}

To generate explanations for the corrective edits, a pilot test was conducted with three models known for their reasoning abilities: Claude 3.7 Sonnet (\url{https://www.anthropic.com/news/claude-3-7-sonnet}), GPT-4o, and Deepseek-R1 (\url{https://huggingface.co/deepseek-ai/DeepSeek-R1}). Qualitative evaluation found that the explanations from all three models were of a similarly high quality (see Appendix \ref{app:explanations} for example explanations from the three models on corrective edits). However, despite its strong performance, Deepseek-R1 was eliminated from consideration due to its significant latency in reasoning, which would negatively impact the user experience in an interactive system. Between the remaining two models, GPT-4o was selected to streamline the technical architecture, as using the same model for both corrections and explanations avoids the overhead of managing a second API call within the surface-level feedback module.

\paragraph{Deep-level LLMs}
For the deep-level feedback module, supervised fine-tuning with the curated dataset of 90 essays relied on same models as the scoring module (LLaMA-3.3-70B and GPT-4o). For LLaMA, the model was fine-tuned using an entropy-based loss function over 8 epochs\footnote{The fine-tuned LLaMA model together with its detailed parameters can be found at \url{https://huggingface.co/judywq/llama-ft-feedback_comment}}. Meanwhile for GPT-4o, the default fine-tuning parameters were unknown due to the black box nature of the commercial model. For direct prompting, aimed at multi-trait feedback inspired by \citet{chu2024}, this research conducted a thorough thematic analysis of the teacher feedback in the training subset. The resulting multiple traits, or structured category codes, are shown in \autoref{tab:feedback_categories}. Based on the two primary categories of macro and micro feedback, the deep-level feedback was further divided into two pipelines. 


\begin{table*}[htbp]
\caption{Categories for deep-level feedback}
\centering

\begin{tabular}{|l|p{0.64\textwidth}|}
\hline
\textbf{Category} & \textbf{Subcategories} \\
\hline
Macro feedback & 
Proper paragraphing; Text length appropriateness; Structure (introduction, body, and conclusion sections) \\
\hline
Micro feedback & 
Context-dependent grammar issues; Clarity of expression; Coherence between ideas; Argumentation quality and logical flow; Formality and academic register \\
\hline
\end{tabular}
\label{tab:feedback_categories}
\end{table*}

For model selection, GPT-4o, LLaMA-3.3-70B-Instruct and Claude 3.7 were adopted among the LLMs this research had utilized to this point. DeepSeek-R1 was excluded due to the long inference time and DeepSeek-V3-671B, due to limited computational resources. Long few-shot prompts were engineered to instruct Claude 3.7 to deliver feedback focusing on these subcategories and offer revision suggestions in both pipelines. The prompts for macro and micro level feedback can be found at Appendix \autoref{deep-level prompts}.

\subsection{Validation scheme}

\subsubsection{Score validation}
For the scoring module, the fine-tuned LLaMA-3.3-70B-Instruct and GPT-4o models were tested using the test subset of 360 essays and the same prompt as the fine-tuning prompt. To evaluate scoring accuracy and reliability, three commonly used metrics were employed: Root Mean Square Error (RMSE), Quadratic Weighted Kappa (QWK), and percentage agreement following \citet{wang2024}, whose results serve as a baseline in this study. RMSE is a measure of accuracy in the form of the absolute difference between system scores and benchmark scores, with lower values indicating better performance \citep{gmd-7-1247-2014}. QWK assessed the level of agreement between system and benchmark scores while considering the ordinal nature of score categories \citep{li-ng-2024-automated}. Percentage agreement is used as a complement to QWK, and we measured the proportion of essays of exact agreement (same score) and adjacent agreement (a difference of 0.5) with benchmark scores.

\subsubsection{Feedback validation}

For surface-level feedback, each original essay and its corresponding corrected version were aligned using ERRANT v3.0.0\footnote{\url{https://github.com/chrisjbryant/errant}} (ERRor ANnotation Toolkit; \citeauthor{bryant-etal-2017-automatic}, \citeyear{bryant-etal-2017-automatic}), a grammar error annotation tool that automatically detects and categorizes all edit operations. Two expert raters evaluated each identified edit along two dimensions:

\begin{itemize}
    \item \textbf{Necessity}: Whether the identified text element genuinely requires correction, given the reported tendency of LLMs to over-correct \citep{katinskaia-yangarber-2024-gpt}.
    \item \textbf{Effectiveness}: Whether the suggested corrective edit effectively addresses the error.
\end{itemize}

The two raters discussed each edit and provided additional comments for cases where edits were deemed unnecessary or ineffective. 

For deep-level feedback, a preliminary check on data integrity and sufficiency of feedback instances was first performed on output from fine-tuned and directly prompted models. The prompt in the macro feedback pipeline asked the LLMs to specifically give feedback to each paragraph, thus the two raters only assessed whether each macro feedback comment was \textit{effective}. For micro feedback,  each was assessed on the dimensions of \textit{necessity} and \textit{effectiveness}, similar to surface-level feedback evaluation. Following the independent evaluations, inter-rater reliability was calculated. For any cases where the two primary raters disagreed, a third expert rater was consulted to deliver the final judgment.

\section{Results}
\subsection{Scoring module}
The RMSE, QWK and percentage agreement between the essay scores generated by the two fine-tuned models and benchmark scores are shown in \autoref{scoring models}. The baseline model is the best performing fine-tuned model in \citet{wang2024}, a prior study that used the same dataset for essay scoring.

\begin{table*}[htbp]
  \caption{Performance metrics of fine-tuned models on the test subset }
  \centering
  \begin{adjustbox}{width=\textwidth}
  \begin{tabular}{lccccc}
    \hline
    \textbf{Fine-tuned Model} & \textbf{RMSE} & \textbf{QWK} & \textbf{Percent (absolute)} & \textbf{Percent (adjacent)} & \textbf{Percent (total)} \\
    \hline
    GPT-4o & 0.44 & 0.84 & 0.45 & 0.48 & 0.93 \\
    LLaMA  & 0.53 & 0.81 & 0.41 & 0.45 & 0.86 \\
    \makecell[l]{Baseline\\ \cite{wang2024}} & 0.57 & 0.78 &0.33&0.52  &0.85\\
    \hline
  \end{tabular}
  \end{adjustbox}
  \label{scoring models}
\end{table*}

Results indicate SOTA performance of our two fine-tuned models in scoring accuracy and reliability, with GPT-4o consistently performing the best across all three evaluation metrics. Specifically, both our fine-tuned models achieved QWK scores above 0.8, surpassing the commonly accepted threshold for near-perfect agreement \citep{sim2005kappa}. For context, ETS considers a QWK of 0.7 to be sufficient for reliable scoring of TOEFL Independent Writing tasks by its \textit{e-rater} system \citep{williamson2012framework}; both of our fine-tuned models exceeded this benchmark. Although ETS does not specify a formal threshold for RMSE, the observed discrepancies of 0.44 (GPT-4o) and 0.53 (LLaMA) from benchmark scores are reasonable, given that ETS permits up to a one-point difference between two human raters \citep{TOEFLscore}, which results in a final averaged score deviating by 0.5 from each individual score.
\subsection{Surface-level feedback module}
After the 40 essays in the test subset were corrected in the surface-level module, ERRANT tagged 2049 edit operations. Human evaluation deemed 1985 edits deemed necessary (96.88\%), out of which 1970 were deemed both necessary and effective (96.14\%). Among the unnecessary edits, two salient patterns have been identified from rater comments. First, GPT-4o seems to prefer British use than American use ($N$=7). For example, it changed \textit{favorite} to \textit{favourite} and moved a period or a comma outside a closing quotation mark (e.g., \textit{``...store.''} to \textit{``...store''.}) to match the British style. The second pattern is related to comma additions ($N$=15). In particular, GPT-4o preferred adding the Oxford comma before the last item in a list of nouns ($N$=10; e.g., \textit{A, B and C} into \textit{A, B, and C}), which was not deemed an error by raters. 



\subsection{Deep-level feedback module}
\subsubsection{Preliminary check}

\begin{sloppypar} 
In terms of output data integrity, the preliminary check showed that neither fine-tuned models generated appropriate output data. For the fine-tuned GPT-4o model, it produced incomplete output for every inference. For each essay, it generated so many comments that the output token exceeded the context window of 8,000 tokens. Thus, each inference was terminated before the complete output was generated. Meanwhile, the fine-tuned LLaMA-3.3-70B model produced formatting errors that prevent its output from being parsed as valid JSON. The most severe issues were structure errors, such as missing key-value delimiters (\texttt{\{"highlighted ", "data":"..."\}} instead of \texttt{\{"highlighted": "...", "data":"..."\}}) and unclosed braces (\texttt{[\{"data": "...", \{"data":"..."\}]} instead of \texttt{[\{"data": "..."\}, \{"data":"..."\}]}). As such, feedback from fine-tuned models were excluded from further human evaluation.
\end{sloppypar}

For directly prompted models, preliminary check on feedback instances revealed that the micro feedback comments from LLaMA-3.3-70B and GPT-4o were insufficient. Of the 40 test essays, LLaMA-3.3-70B generated 336 comments and GPT-4o generated 368 comments, the averages of both were below 10. Claude 3.7, however, generated 628 comments. To verify that this was a difference in quality and not just verbosity, the researchers conducted a qualitative review of three randomly selected essays. This comparison confirmed that the additional comments from Claude 3.7 were indeed more fine-grained and pedagogically necessary. Given the intensive human labor required for a full expert evaluation, a decision was made to focus validation efforts exclusively on the feedback from the most promising model, Claude 3.7.

\subsubsection{Macro feedback}
The 40 essays in the test subset contained 201 paragraphs and thus Claude 3.7 generated 201 macro feedback comments. The contingency table between the two raters on the effectiveness of the comments is shown in \autoref{macro contigency}.

\begin{table}[htbp]
\caption{Contingency table of macro comment evaluation by two raters}
\centering

\begin{tabular}{lcc}
\toprule
\textbf{Rater A / Rater B} & \textbf{Effective} & \textbf{Ineffective} \\
\midrule
\textbf{Effective} & 179 & 9 \\
\textbf{Ineffective}   & 13  & 0 \\
\bottomrule
\end{tabular}
\label{macro contigency}
\end{table}

To assess inter-rater reliability, Gwet's $AC1$ coefficient \citep{gwet2008computing} was employed, which ranges from $-1$ to $+1$ similar to Cohen's Kappa. This measure was selected over Cohen's Kappa due to its robustness against the ``Kappa paradox'', where high observed agreement coincides with low kappa values in cases of skewed marginal distributions (i.e., the dominance of the \textit{effective} category in rater evaluation) \citep{wongpakaran2013comparison, gwet2008computing}. The $AC1$ coefficient was calculated to be 0.89, with a standard error ($SE$) of 0.03 and a 95\% confidence interval ($CI$) ranging from 0.82 to 0.93. As there is no commonly accepted interpretation of Gwet's $AC1$ coefficient, we relied on the interpretation of Cohen's Kappa to determine the degree of agreement \citep{landis1977measurement}. The results thus indicate very strong and statistically significant agreement beyond chance ($p <$ .001).

After a third rater resolved the disagreement of the two raters, the number of effective comments was finalized as 187 (93.03\%) and that of ineffective comments, 27 (6.97\%). For the 27 ineffective comments, remarks from raters revealed that Claude 3.7 was not flexible enough to handle unanticipated input variations. For example, one test-taker included a title in their essay, a component not required in the writing task, and Claude 3.7 misinterpreted it as the first paragraph. In another case, some test-takers improperly formatted each sentence as a separate paragraph, which led to repeated comments that a paragraph was too short and thus should be combine with the previous one ($N$=8). In addition, some test-takers were unable to complete the task within the allotted time. Claude 3.7 failed to recognize the time constraint of such unfinished responses and instead provided detailed guidance on composing conclusions ($N$=2). Also very interesting is that Claude 3.7 was not aware of the test-taker situation where they didn't have access to external resources and recommended incorporating research evidence and citations to strengthen arguments ($N$=1). 

\subsubsection{Micro feedback}
For the same 40 essays, Claude 3.7 generated 630 micro feedback comments. The contingency table between the two raters on the necessity of the comments is show in \autoref{necessity contigency}, while that on the effectiveness of necessary comments mutually agreed by two raters is shown in \autoref{effectiveness contingency}.

\begin{table}[htbp]
\caption{Contingency table of necessity of micro feedback by two raters}
\centering

\begin{tabular}{lcc}
\toprule
\textbf{Rater A / Rater B} & \textbf{Necessary} & \textbf{Unnecessary} \\
\midrule
\textbf{Necessary} & 579 & 28 \\
\textbf{Unnecessary}  & 22  & 1 \\
\bottomrule
\end{tabular}
\label{necessity contigency}
\end{table}

\begin{table}[ht]
\caption{Contingency table of effectiveness judgment for micro feedback by two raters}
\centering

\begin{tabular}{lcc}
\toprule
\textbf{Rater A / Rater B} & \textbf{Effective} & \textbf{Ineffective} \\
\midrule
\textbf{Effective} & 577 & 1 \\
\textbf{Ineffective}  & 1  & 0 \\
\bottomrule
\end{tabular}
\label{effectiveness contingency}
\end{table}

Given the skewed distribution of judgments in both dimensions, Gwet's $AC1$ \citep{gwet2008computing} was again employed. The resulting agreement on necessity was very strong and statistically significant beyond chance ($AC1$= 0.91; $SE$=0.01; 95\% $CI$, [0.89, 0.94]; $p <$ .001). The resulting agreement on effectiveness was also very strong and statistically significant beyond chance ($AC1$= 1.00; $SE$=0.00; 95\% $CI$, [0.99, 1.00]; $p <$ .001).

After a third rater resolved the disagreement between the two raters, the number of necessary comments was finalized as 600 (95.24\%), that of unnecessary ones, 30 (4.76\%). Among the necessary comments, 596 were deemed effective while 4 were ineffective, and thus the proportion of both necessary and effective comments is 94.69\%.

Based on rater comments, one salient pattern was  observed: some micro comments overlapped with macro ones. Claude 3.7 at times expand the comment for a sentence to cover the whole paragraph ($N$=18). For instance, when the first sentence in an introduction paragraph did not properly set the background of the topic, Claude 3.7 would comment that there was a lack of proper background and then went on to give more feedback on how to write a good introduction paragraph.

\subsection{Final system architecture and UI design}

\begin{figure}[htbp]
    \centering
    \includegraphics[width=1\linewidth]{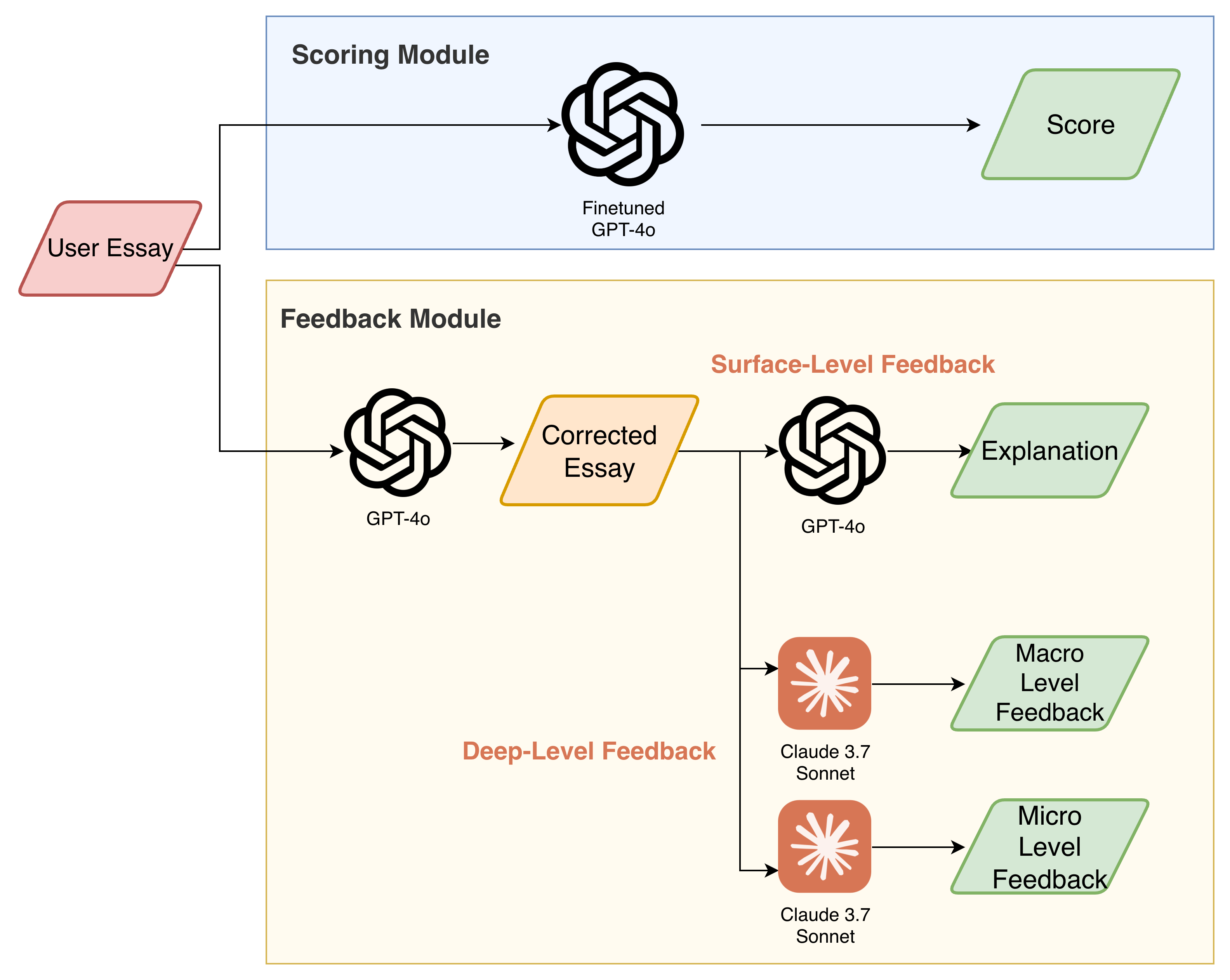}
    \caption{Final system architecture}
    \label{system architecture}
\end{figure}

Based on the evaluation results, the system architecture was finalized by selecting LLMs tailored to each module: the fine-tuned GPT-4o model for the scoring module; direct prompting with GPT-4o for the surface-level feedback module; and Claude 3.7 for both pipelines in the deep-level feedback module. \autoref{system architecture} illustrates the overall workflow of our system.

\begin{figure}[t]
    \centering
    \includegraphics[width=0.7\linewidth]{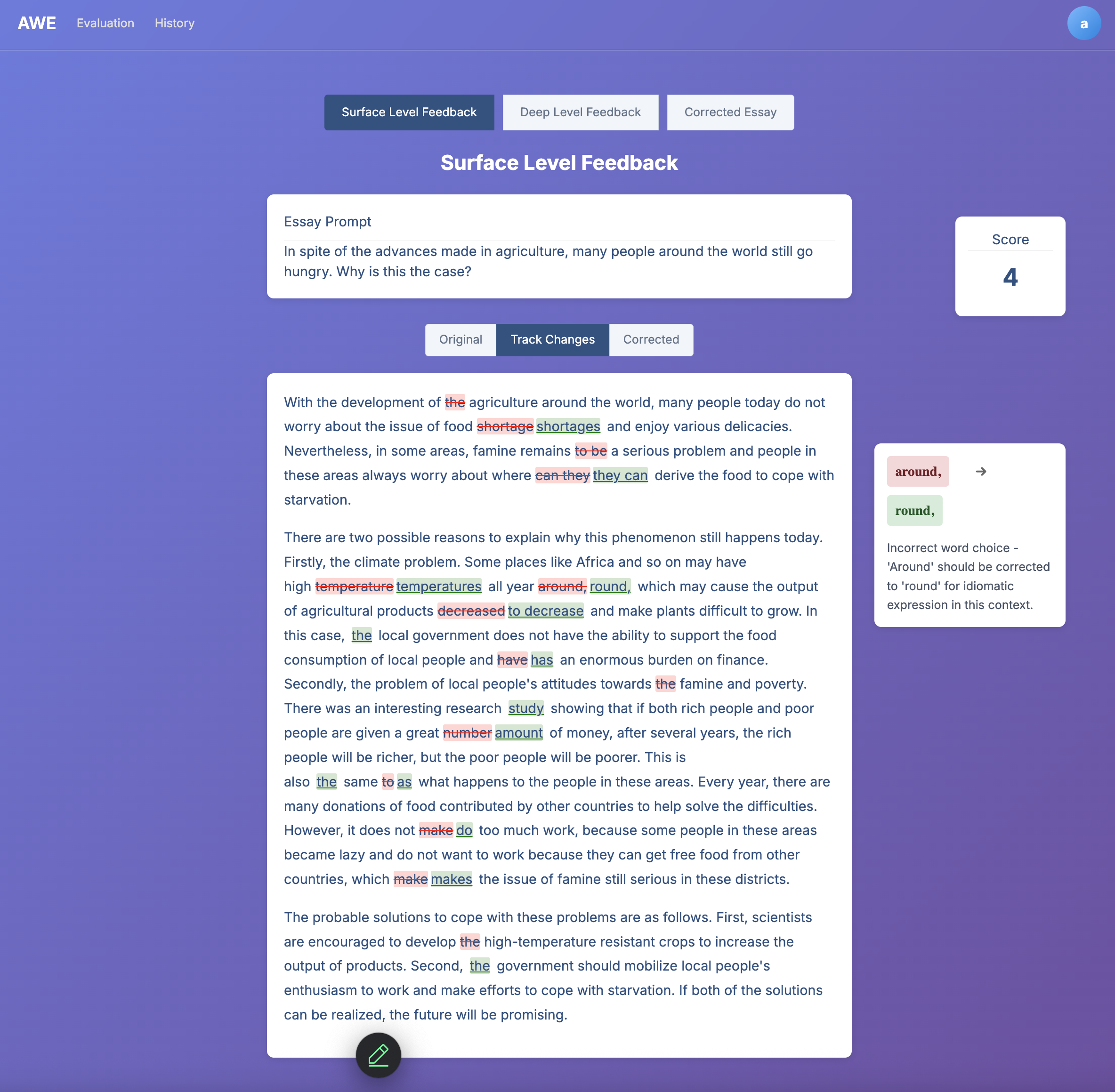}
    \caption{Screenshot of the surface-level feedback page - track changes}
    \label{surface level}
\end{figure}

\begin{figure}[t]
    \centering
    \includegraphics[width=0.7\linewidth]{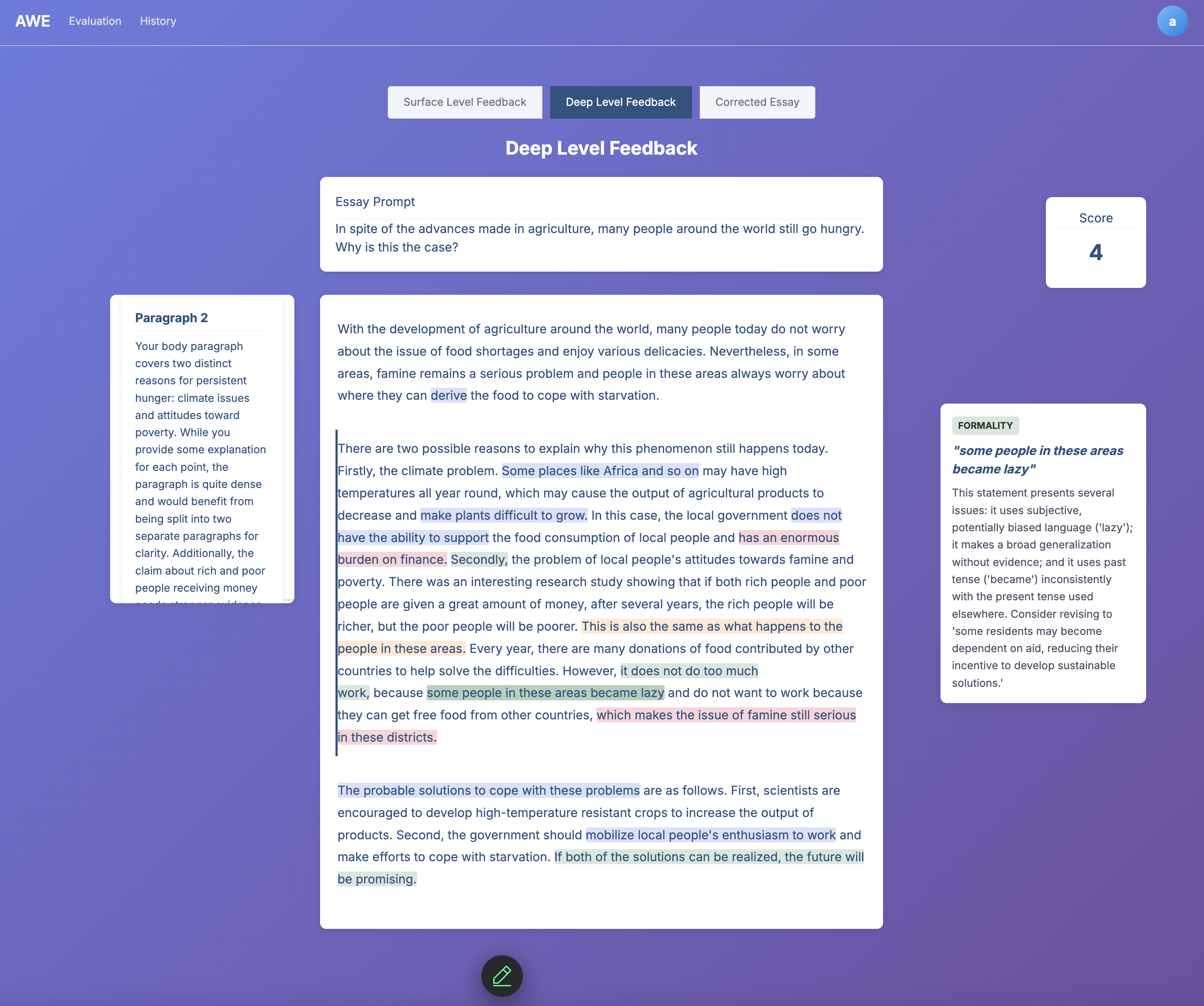}
    \caption{Screenshot of the deep-level feedback page}
    \label{deep level}
\end{figure}

To mimick human feedback mechanisms, an interactive UI was developed. 
The feedback page shows the predicted score of an essay and includes a surface-level feedback tab and a deep-level feedback tab. The surface-level feedback tab allows three display modes of surface-level feedback: (1) the original writing with erroneous elements highlighted, (2) a combination of the original and corrected writing with edit operations shown, and (3) the corrected writing with corrections highlighted (see~\autoref{surface level}). For the deep-level feedback tab, a comment-based interface design was adopted, modeled after Microsoft Word's Comment feature. Macro-level comments are displayed in a left-side panel and are aligned with paragraph boundaries. Micro issues under coherence, clarity, grammar, argumentation and formality are highlighted within the essay text using color-coded categories. When a user hovers the mouse over a highlighted section, the corresponding feedback comment appears on the right-side panel (see~\autoref{deep level}).


\section{Discussions}
The AWE system built on a modular approach in this research successfully delivered accurate and reliable scores benchmarked against ETS standards for TOEFL independent writing and high-quality feedback validated by human teachers. The findings further support previous research by \citet{stahl2024} and \citet{chu2024}, who have demonstrated that specialized models handling different aspects of writing evaluation can produce better assessments than unified approaches. 

Human evaluation of surface-level feedback shows that GEC tasks can be addressed effectively by well-designed prompts without the need for sophisticated fine-tuning. This finding echoes the study by \citet{zeng-etal-2024-evaluating}, who demonstrated that LLMs with appropriate prompting can achieve competitive performance on GEC tasks compared to specialized fine-tuned models. Particularly, we successfully minimized over-correction through our targeted prompt, an issue previously identified as a key limitation of LLM-prompting-based GEC approach \citep{katinskaia-yangarber-2024-gpt,davis-etal-2024-prompting}. The high precision rate of 96.14\% indicates that when the system identifies errors, it is nearly always correct in doing so. 

Human validation of deep-level feedback suggests that well-structured few-shot prompting can generate effective and structured comments on higher-order thinking, and may even be more effective than models of supervised fine-tuning. Here the researchers would like to focus on why supervised fine-tuning did not yield satisfactory results on deep-level feedback. The fine-tuned GPT model that provided feedback to almost every word, phrase, and sentence, most of which were unnecessary. As the fine-tuning parameters were unknown, it can only be speculated that supervised fine-tuning with non-deterministic output is not well-suited for GPT-4o. Contrary to numerical and categorical output, there are no clear standards as to whether a comment is necessary and/or effective. Further reinforcement learning from human feedback (RLHF) may be needed to guide the fine-tuned GPT-4o to understand what feedback is necessary and preferred. For the fine-tuned LLaMA model, despite the data parsing issues, it was found that the feedback comments were actually quite similar to those in the fine-tuning subset, though still not so good as those from Claude 3.7. This indicates that the supervised fine-tuning did enhance model performance in generating feedback. Data formatting issue is likely an inherent issue with LLaMA, as GPT-4o did not exhibit such an issue. A previous study by \citet{yao2025reff} specifically highlighted the limitations of LLaMA in data formatting tasks, which prompted the researchers to develop a format benchmark to evaluate and improve formatting capabilities of LLMs. If the data formatting issue can be addressed, perhaps by incorporating a data checker tool, it may be possible to effectively use fine-tuned LLaMA and potentially other open-source models for generating structured feedback. 
\paragraph{Pedagogical implications}
With an interactive UI, the AWE system has strong pedagogical implications. The system can immediately benefit students engaged in argumentative writing. They can receive prompt feedback not available in traditional classroom settings, where it usually takes teachers days to provide comprehensive feedback on student writing. Teachers can also use this tool to alleviate their workload, particularly for surface-level feedback. Instead of spending time correcting basic grammatical and mechanical errors, they can focus more on guiding students' higher-order thinking in argumentative writing, aided by the feedback generated by the system. At the institutional level, the system can also serve as a reference for scoring to ensure fairness. Particularly for large-scale coordinated writing courses taught by various instructors, it is common for teachers to develop idiosyncratic scoring criteria. Having an automated reference score could promoting more consistent evaluation standards.

\section{Limitations}

There are a few limitations in the design of this study. First, the system was developed using a proprietary dataset that has only been used in one prior AWE study. This limits the scope of direct comparisons on scoring performance. Second, in the validation scheme, a key limitation is that human evaluation only focused on precision without considering recall. For both surface-level and deep-level feedback, the necessity and effectiveness of feedback were assessed but it was not evaluated whether the system captured all errors that should have been identified or all text elements that should have required feedback comments. A more robust evaluation would require establishing a gold standard of all possible errors and feedback opportunities against which system performance could be measured. Unfortunately, the extensive human resources required to achieve this was beyond our capacity. In addition, there is a lack of measurement of actual impact on student learning outcomes, which was beyond the scope of this research. Future research should include experimental studies tracking writing improvement over time when using the system compared to traditional feedback methods.

There are limitations with the system as well. First, apart from the salient issues identified through rater comments, instances of LLM hallucinations were also observed. For example, there were cases, though infrequent, where the system identified non-existent errors and subsequently suggested ``corrections'' that were identical to the original text. Future iterations should incorporate verification mechanisms, such as adding another LLM as the reviewer of generated feedback. Second, the system is limited in its generalizability, as it was developed and tested exclusively on argumentative essays with writing prompts. It cannot be directly applied to other types of argumentative writing, such as source-based writing where writers are required to incorporate or respond to source information. Third, the feedback offered by the system is solely in English, creating barriers for learners who might struggle to comprehend feedback comments. Future iterations should incorporate multilingual feedback capabilities, including translation of comments into learners' first languages to facilitate comprehension and implementation of suggestions.

From a practical implementation perspective, there are limitations arising from the reliance on commercial API services. There are potential scalability issues related to cost, rate limits, and long-term sustainability. In addition, constant changes to commercial models may impact system performance. Data privacy concerns also emerge when student writing is processed through third-party services.






\section{Conclusions}
This research developed and validated \textit{WrAFT}, a modular LLM-powered automated writing evaluation system designed specifically for argumentative essays. By decomposing AWE tasks into scoring, surface-level feedback, and deep-level feedback modules, the system achieved state-of-the-art performance in scoring on a proprietary TOEFL writing dataset. Human validation of system-generated feedback also suggest satisfactory results. The findings of this research provide a strong foundation for future research in modular AWE systems that provide both scoring and comprehensive feedback. An interactive UI has also been released for public use free of charge. 

\bibliographystyle{elsarticle-harv}
\bibliography{bibliography}
\clearpage







\begin{appendices}

\section{Explanations of surface-level corrective edits by GPT-4o, Claude 3.7 and DeepSeek-R1}
\label{app:explanations}

\begin{table}[H]
\caption{Explanations of corrective edits generated by Claude 3.7 Sonnet, GPT-4o, and DeepSeek-R1.}
\centering
\footnotesize
\begin{adjustbox}{max width=\linewidth,max totalheight=\textheight,keepaspectratio}

\begin{tabular}{c p{3cm} p{5cm} p{4.5cm} p{4.5cm}}
\toprule
\textbf{ID} & \textbf{Original / Correction} & \textbf{Claude 3.7 Sonnet} & \textbf{GPT-4o} & \textbf{DeepSeek-R1} \\
\midrule
0 & \textit{serve \textrightarrow{} serves} & Subject--verb agreement error. ``Academic writing'' is singular and requires ``serves''. & Subject--verb agreement error. ``Writing'' is singular, so verb should be ``serves''. & Subject--verb agreement error. ``Academic writing'' is singular and requires ``serves''. \\
1 & \textit{corner \textrightarrow{} cornerstone} & Word choice improvement. ``Cornerstone'' conveys the foundation or essential component of scholarly communication. & ``Corner'' is semantically incorrect; ``cornerstone'' is the correct metaphor for something fundamental. & ``Corner'' is incomplete; ``cornerstone'' better conveys fundamental importance. \\
2 & \textit{enable \textrightarrow{} enables} & Subject--verb agreement. ``It'' is singular and requires ``enables''. & Subject--verb agreement. ``It'' is singular, so verb should be ``enables''. & Subject--verb agreement. ``It'' is singular and requires ``enables''. \\
3 & \textit{they're \textrightarrow{} their} & Incorrect homophone. ``They're'' (they are) should be replaced with possessive ``their''. & Incorrect homophone. ``They're'' is incorrect; should be possessive ``their''. & Wrong word choice. ``They're'' (they are) vs. ``their'' (possessive). \\
4 & \textit{accurate \textrightarrow{} accurately} & Incorrect part of speech. Adverb ``accurately'' modifies verb ``convey''. & Adverb needed. ``Accurate'' \textrightarrow{} ``accurately''. & Adverb required. ``Accurate'' should be ``accurately'' to modify verb. \\
5 & \textit{spends \textrightarrow{} spend} & Subject--verb agreement. Plural ``Scholars'' requires plural ``spend''. & Subject--verb agreement. ``Scholars'' is plural \textrightarrow{} ``spend''. & Subject--verb agreement. ``Scholars'' plural \textrightarrow{} ``spend''. \\
6 & \textit{hone \textrightarrow{} honing} & Verb form consistency. ``Honing'' maintains parallel structure with ``learning''. & Verb form error. Change to gerund ``honing'' for parallelism. & Gerund required after ``years''. ``Honing'' shows ongoing action. \\
7 & \textit{learn to balance technic \textrightarrow{} learning to balance technical} & Verb form consistency \& spelling. ``Learning'' parallels ``honing''; ``technic'' \textrightarrow{} ``technical''. & Grammar and clarity. ``Learn to balance technic'' \textrightarrow{} ``learning to balance technical''. & Parallel structure and word correction. ``Learning'' matches ``honing''; ``technical'' corrects ``technic''. \\
8 & \textit{discipline \textrightarrow{} disciplines} & Pluralization. Remove comma; use plural ``disciplines''. & Number agreement. ``Many'' \textrightarrow{} plural ``disciplines''. & Plural required. ``Many'' \textrightarrow{} plural ``disciplines''. \\
9 & \textit{help \textrightarrow{} helps} & Subject--verb agreement. ``Process'' is singular \textrightarrow{} ``helps''. & Subject--verb agreement. ``The peer review process'' is singular \textrightarrow{} ``helps''. & Subject--verb agreement. ``Process'' is singular \textrightarrow{} ``helps''. \\
10 & \textit{distribute \textrightarrow{} distributed} & Verb form. Passive voice requires past participle ``distributed''. & Verb tense. Use past participle ``distributed'' in passive construction. & Passive voice requires past participle. ``Is distributed'' correct form. \\
\bottomrule
\end{tabular}
\end{adjustbox}
\label{tab:explanation_comparison}
\end{table}

\section{Fine-tuning and inference prompt for the scoring module}
\label{app:scoring prompt}
\begin{Verbatim}[fontsize=\small]
<instructions>
As a language expert, your task is to evaluate 
argumentative essays on a scale of 0 to 5 
(with 0.5 increments) based on the rubrics below.

<rubric>
[Complete rubrics here]
</rubric>
</instructions>

<input_data_structure>
{{
  "essay_prompt": "...", // The essay prompt
  "essay_text": "...", // The essay text
}}
</input_data_structure>

<output_data_structure>
{{
  "score": 0.5 // The score
}}
</output_data_structure>

Here's the input data:
<input_data>
{{
  "essay_prompt": {essay_prompt},
  "essay_text": {essay_text}
}}
</input_data>
\end{Verbatim}

\section{Surface-level prompt}
\label{app:surface-level prompt}
\begin{Verbatim}[fontsize=\small]
<instructions>
You are an English linguist and your task is to correct the grammatical 
and mechanical errors in an English essay. Do not alter word choices 
unnecessarily (e.g., replacing words with synonyms) or make stylistic 
improvements. Keep the original paragraphing and DO NOT remove or add 
any paragraphs.

Requirements:
1. The output should be in JSON format.
2. The output should be in the same format as the output_data_structure.
</instructions>

<output_data_structure>
{{
  "corrected_paragraphs": [
    "...", // The corrected text of the first paragraph
    "...", // The corrected text of the second paragraph
    ...
  ]
}}
</output_data_structure>

Here's the input data:
<input_data>
{{
  "essay_paragraphs": {essay_paragraphs}
}}
</input_data>
\end{Verbatim}


\section{Deep-level prompts}
\label{deep-level prompts}
\subsection{Prompt for macro feedback}

Instructions:

You are an English teacher in argumentative writing and your job is to give macro feedback to a student's argumentative essay structure. An essay should typically include three major sections: Introduction, Body and Conclusion. 

The Introduction part should open with some background of the essay topic, followed by a thesis statement of the author and possibly end with a predictor of sub-points for later paragraphs.

The body part should contain detailed argumentation of sub-points of the thesis statement, probably divided into several paragraphs, with each paragraph having a clear topic sentence. 

The conclusion part should efficient summaries the essay and preferably end with a forward looking statement. 

You need to first identify the three sections, and then give feedback to each paragraph. Specifically, you need to identify the issues of essay structure targeting specific paragraph(s). 
\begin{itemize}
    \item In the introduction paragraph, there may be a lack of background information, thesis statement (main claim) or a predictor for following content, or the introduction does not effectively introduce the topic; 
    \item In body paragraphs, there may be a lack of topic sentences and issues in argumentation. Specifically for argumentation, focus on ideas that are irrelevant to the thesis statement or the claim; repeated and thus redundant ideas, underdeveloped arguments where there is a lack of evidence and support, e.g, in the form of examples, or the support is poor or irrelevant; where there is a lack of warrant; where the author's stance is confusing; and where handling of counter-argument is poor, etc. 
    \item In the conclusion paragraph, there may be undiscussed ideas, ideas that shouldn't be placed in the conclusion, such as further arguments or counter-arguments, and lack of effective summary of main points, etc. 
    \item For all the three section, please consider whether proper paragraphing is there. Improper paragraphing may include putting content that should be in the body part, e.g., development of main points, examples, counter-arguments, etc., with the introduction or conclusion part, and vice versa.
\end{itemize}

For each paragraph, write down feedback comments, including suggestions for solutions, possibly with an example of how to rewrite. Please make sure the number of comments your return is equal to the number of paragraphs and use the second person pronoun "you" to sound like you are talking to the student directly. Do not specifically divide your comments into "issue:" and "solution:" points; instead write it as a coherent piece of text.

Input and Output

You will be given the essay prompt, the essay text of the student in the user input.
Your response should be in the following JSON format:
\begin{verbatim}
    {
  "comments": [
    {
      "id": 0, // The id of the comment
      "Paragraph": , // The number of the specific paragraph
      "data": "...", // The teacher's feedback comment
    },
    ...
  ]
}

\end{verbatim}
\subsection{Prompt for micro feedback}
You are an English teacher specializing in argumentative writing. Your task is to provide detailed micro feedback on a student's argumentative essay, focusing on grammar, clarity, formality, and coherence. You will analyze the essay and provide specific, actionable feedback for improvement.

\begin{verbatim}
Here is the essay prompt:
<essay_prompt>
{{essay_prompt}}
</essay_prompt>

And here is the student's essay:
<essay_text>
{{essay_text}}
</essay_text>

\end{verbatim}

Please carefully read and analyze the essay, then provide feedback according to the following guidelines:

1. Focus Areas:

   a) Grammar:
   
      - Identify and correct grammatical and mechanical errors
      
      - Address context-dependent grammar issues (e.g., appropriate tense, aspect)
      
   b) Clarity:
   
      - Highlight unclear, inaccurate, or redundant meanings
      
      - Point out lengthy or confusing sentence structures
      
      - Identify poor or unclear referents
   c) Formality:
   
      - Note expressions requiring hedging
      
      - Identify informal language use (e.g., coordinating conjunctions at sentence beginnings, subjective wording, informal interrogatives, colloquial language, second-person pronouns, contractions)
      
   d) Coherence:
   
      - Evaluate connections between sentences and ideas
      
      - Assess transitions between paragraphs
      
      - Check consistency in person perspective

2. Feedback Structure:

   For each issue you identify, provide the following information:
   
   - The type of issue (grammar, clarity, formality, or coherence)
   
   - The specific paragraph number (starting from 0) and the micro text span (preferably words or phrases, avoiding clauses and sentences when possible)
   
   - A clear explanation of the issue
   
   - A suggested solution

3. Additional Guidelines:

   - If similar issues occur multiple times, acknowledge this in your feedback (e.g., "As mentioned previously...")
   
   - Address the student directly using the second-person pronoun "you"
   
   - Write your feedback as coherent paragraphs rather than dividing into explicit "issue" and "solution" sections
   
   - Ensure your comments are not repetitive
   
   - Highlight the shortest text spans possible for feedback, unless a longer span is absolutely necessary

4. Output Format:

   Your final output should be in JSON format, structured as follows:
   \begin{verbatim}
       
   {
     "comments": [
    {
      "id": 0, // The id of the comment
	   "Paragraph": , // The number of the specific paragraph, 
                     // starting from 0
      "start": 22, // The start character position of the 
                   // highlighted text in the paragraph
      "end": 36, // The end character position of the 
                 // highlighted text in the paragraph
      "highlighted_text": "...", // The highlighted text 
                                // span to be commented on
      "data": "...", // The teacher's feedback comment 
                    // (explanation of the issue and solution)
      "type": "..." // The type of issue, one of "grammar",
                    // "clarity", "formality", "coherence"},
       ...// More comments...
     ]
   }
 \end{verbatim}
Before providing your final JSON output, wrap your analysis inside \path{<essay_analysis>} tags in your thinking block. Follow these steps:

1. Grammar analysis:

   - List specific grammatical errors found
   
   - Note any context-dependent grammar issues

2. Clarity analysis:

   - Identify unclear or redundant phrases
   
   - List confusing sentence structures
   
   - Note any poor referents
   
3. Formality analysis:

   - List instances of informal language use
   
   - Identify expressions that require hedging

4. Coherence analysis:

   - Evaluate connections between sentences and paragraphs
   
   - Note any inconsistencies in perspective
   
   - Avoid commenting on general essay structure (general comments on introduction, body and conclusion paragraphs)

For each issue, consider multiple perspectives and potential solutions. This thorough analysis will ensure accurate and helpful feedback.

Begin your analysis now, followed by the JSON-formatted feedback. Your final output should consist only of the JSON-formatted feedback and should not duplicate or rehash any of the work you did in the thinking block.






\end{appendices}


\end{document}